\documentclass[aps,pra,superscriptaddress,twocolumn,longbibliography]{revtex4-1}
\usepackage{mathrsfs}
\usepackage{amsfonts}
\usepackage{amssymb}
\usepackage{amsmath}
\usepackage{graphicx}
\usepackage{lipsum} 
\usepackage{ulem}
\usepackage{sidecap}
\usepackage{color}
\usepackage{tikz}
\usepackage[colorlinks=true, urlcolor=blue, citecolor=blue,linkcolor=blue,citebordercolor={1 0 0},linkbordercolor={0 0 1}]{hyperref}
\usepackage{subeqnarray}
\usepackage{verbatim}
\usepackage{euscript} %curly fonts mathcal and mathscr
\usepackage[columnwise]{lineno}

\usepackage{array}
\usepackage{multirow}

\renewcommand{\eqref}[1]{Eq.~(\ref{#1})} % Reference to equation
\definecolor{mydarkblue}{rgb}{0,0.08,0.45}
\definecolor{myfavblue}{rgb}{0.1176, 0.392, 1.0}
\hypersetup{
    colorlinks=true,
    linkcolor=mydarkblue,
    citecolor=mydarkblue,
    filecolor=mydarkblue,
    urlcolor=mydarkblue}

\newcommand{\utchem}{Department  of  Chemistry,  University  of  Toronto,  Toronto,  Ontario  M5G 1Z8,  Canada}
\newcommand{\utcomp}{Department  of  Computer Science,  University  of  Toronto,  Toronto,  Ontario  M5S 2E4,  Canada}
\newcommand{\vectorinst}{Vector  Institute  for  Artificial  Intelligence,  Toronto,  Ontario  M5S  1M1,  Canada}
\newcommand{\cifar}{Canadian  Institute  for  Advanced  Research,  Toronto,  Ontario  M5G  1Z8,  Canada}

\begin{document}

\title{Language models can generate molecules, materials, and protein binding sites directly in three dimensions as XYZ, CIF, and PDB files}

\author{Daniel Flam-Shepherd}
%\email{danielfs@cs.toronto.edu}
\affiliation{\utcomp}
\affiliation{\vectorinst}

\author{Al\'an Aspuru-Guzik}
%\email{alan@aspuru.com}
\affiliation{\utcomp}
\affiliation{\vectorinst}
\affiliation{\utchem}
\affiliation{\cifar}

\begin{abstract}
Language models are powerful tools for molecular design. Currently, the dominant paradigm is to parse molecular graphs into linear string representations that can easily be trained on. This approach has been very successful, however, it is limited to chemical structures that can be completely represented by a graph-- like organic molecules-- while materials and biomolecular structures like protein binding sites require a more complete representation that includes the relative positioning of their atoms in space. 
In this work, we show how language models, without any architecture modifications, trained using next-token prediction-- can generate novel and valid structures in three dimensions from various substantially different distributions of chemical structures.
In particular, we demonstrate that language models trained directly on sequences derived directly from chemical file formats like XYZ files, Crystallographic Information files (CIFs), or Protein Data Bank files (PDBs) can directly generate molecules, crystals, and protein binding sites in three dimensions. 
Furthermore, despite being trained on chemical file sequences-- language models still achieve performance comparable to state-of-the-art models that use graph and graph-derived string representations, as well as other domain-specific 3D generative models.  
In doing so, we demonstrate that it is not necessary to use simplified molecular representations to train chemical language models-- that they are powerful generative models capable of directly exploring chemical space in three dimensions for very different structures. 
\end{abstract}

\maketitle

%\linenumbers

\begin{figure*}[t]
\centering
\includegraphics[width=\linewidth]{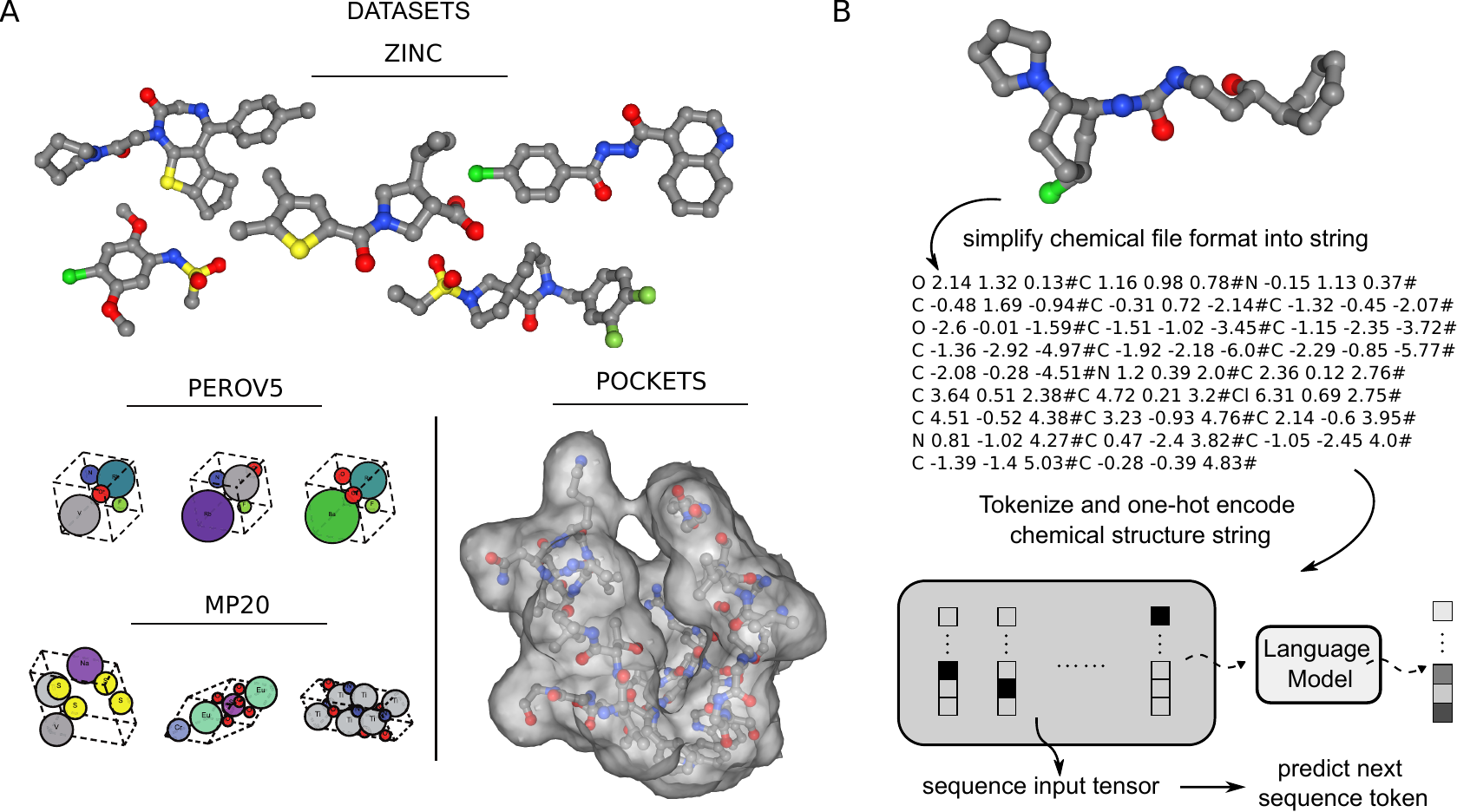}
\caption{A) The training datasets of structures that we benchmark language models on in this work. B) The overview of the training workflow -- chemical file formats are converted to sequences of tokens using either character or coordinate-level tokenization. The language model is trained to predict the next token in these sequences.}
\label{fig:f1}
\end{figure*}

\section{Introduction}

Language models are autoregressive models for sequence generation that have shown impressive progress recently in natural language understanding using deep neural networks  \cite{brown2020language,radford2018improving,radford2019language}. These advancements are driven by architecture improvements like the Transformer-- a powerful neural network for sequential data that uses self-attention \cite{vaswani2017attention}. Transformers have found use in many significant scientific applications like protein structure prediction and design \citep{jumper2021highly,ingraham2019generative,jin2021iterative} and various tasks in cheminformatics \cite{schwaller2019molecular, fuchs2020se, zhou2022uni}. An important scientific objective is the exploration of chemical space, in order to discover new drugs and materials \cite{sanchez2018inverse}. Language models have enormous potential for chemical space exploration-- which remains almost entirely unexplored given the $10^{60}$ drug-like molecules \cite{Polishchuk2013} and even more potentially accessible materials. 
Already, recent large language models \cite{brown2020language, openai2023gpt4} are having an impact on research in chemistry and molecular design \cite{white2023assessment,bran2023chemcrow,boiko2023emergent}.

Substantial work has been done using other neural networks in the exploration of chemical space-- deep generative models \citep{Gomez-Bombarelli2016, jin2018junction, jin2020hierarchical, FlamShepherd_2021} can be trained on large datasets to generate novel functional compounds from any target distributions.
An important question arises on how to best represent a molecule when training models to learn molecular representations. There are many different model architectures appropriate for different representations, indeed the task at hand will heavily impact design choice. 
A popular approach is to directly use molecular graphs and make use of geometric deep learning to learn representations directly on atoms and bonds \citep{DuvMacetal15nfp, flam2021neural,li2018learning,liu2018constrained,jin2018junction, you2018graph,seff2019discrete}-- 
this strategy has been used to screen large compound libraries-- one attempt lead to the discovery of novel antibiotics \citealp{stokes2020deep}. 

%string representations
An alternate approach is to use SMILES or SELFIES string representations \citep{weininger1988smiles, krenn2019selfies} that linearize molecular graphs into strings. SMILES and SELFIES are widely used for machine learning-assisted molecular design \citep{gomez2018automatic,segler2018generating, kusner2017grammar}. SMILES and SELFIES are convenient for neural networks designed for sequences-- which have proven to be powerful generative models of natural language \cite{sutskever2011generating, brown2020language}. Indeed, they have been used to achieve state-of-the-art results with chemical language models using Long Short Term Memory networks (LSTMs) \cite{flam2022language, hochreiter1997long}. 

% problems with graphs and strings
However, strings and graphs are a simplified representation of molecules--  which are at naturally represented as point clouds of atoms that includes their three-dimensional (3D) positions in space. For many molecular design tasks, such as catalysis \cite{chanussot2021open}, this geometric information is essential. SMILES and SELFIES sidestep this complete representation of molecules and are entirely unable to represent materials-- which cannot be simplified as graphs and have a complex periodic 3D structure. Indeed, the geometric structure of molecules and materials is an important determinant of their properties. Indeed, molecules and materials are complex structured data-- discrete in terms of their atomic elements but continuous in terms of the coordinates of those elements.

Any 3D molecule, biomolecule, or material can be fully represented and stored as text data in XYZ, PDB, or CIF  files (as the most common formats). These text files are effectively long strings defined by atom coordinate pairs and other information-- to directly model these files as strings using a language model is somewhat unintuitive given the continuous nature of 3D space. Indeed, for success, a language model has to learn many layers of validity-- from the basic elements of the file structure to the complex spatial arrangements of atoms in any molecule or material. Given the impressive ability of language models to model complex molecular distributions using simple string representations \cite{flam2022language}-- is it possible for language models to generate molecules and crystals in three dimensions by training on entire XYZ, CIF or even PDB files? There are many structural distributions of molecules and materials that simple string representations cannot model and many potential domain-specific applications that can be tackled-- if language models could directly model more complex chemical file formats.   

In recent work, a few models have achieved state-of-the-art results by focusing on generating molecules and materials in 3D in a way that satisfies permutation, translation, rotation, and periodic invariances with SE(3) equivariant architectures \cite{hoogeboom2022equivariant, xiecrystal, gebauer2019symmetry, garcia2021n}. In contrast, for a language model that has none of these invariances built in-- it is challenging to generate structures by placing atoms using absolute or Cartesian coordinates. Chemical format files can easily turn into very long sequences, even for small molecules-- chemical language models using LSTMs will inevitably have issues learning important long-range dependencies. However, other architectures like Transformers process the entire sequence at a time and do not have this issue-- and therefore are the model that is most likely to succeed at this task.

\begin{figure*}[t]
\centering
\includegraphics[width=\linewidth]{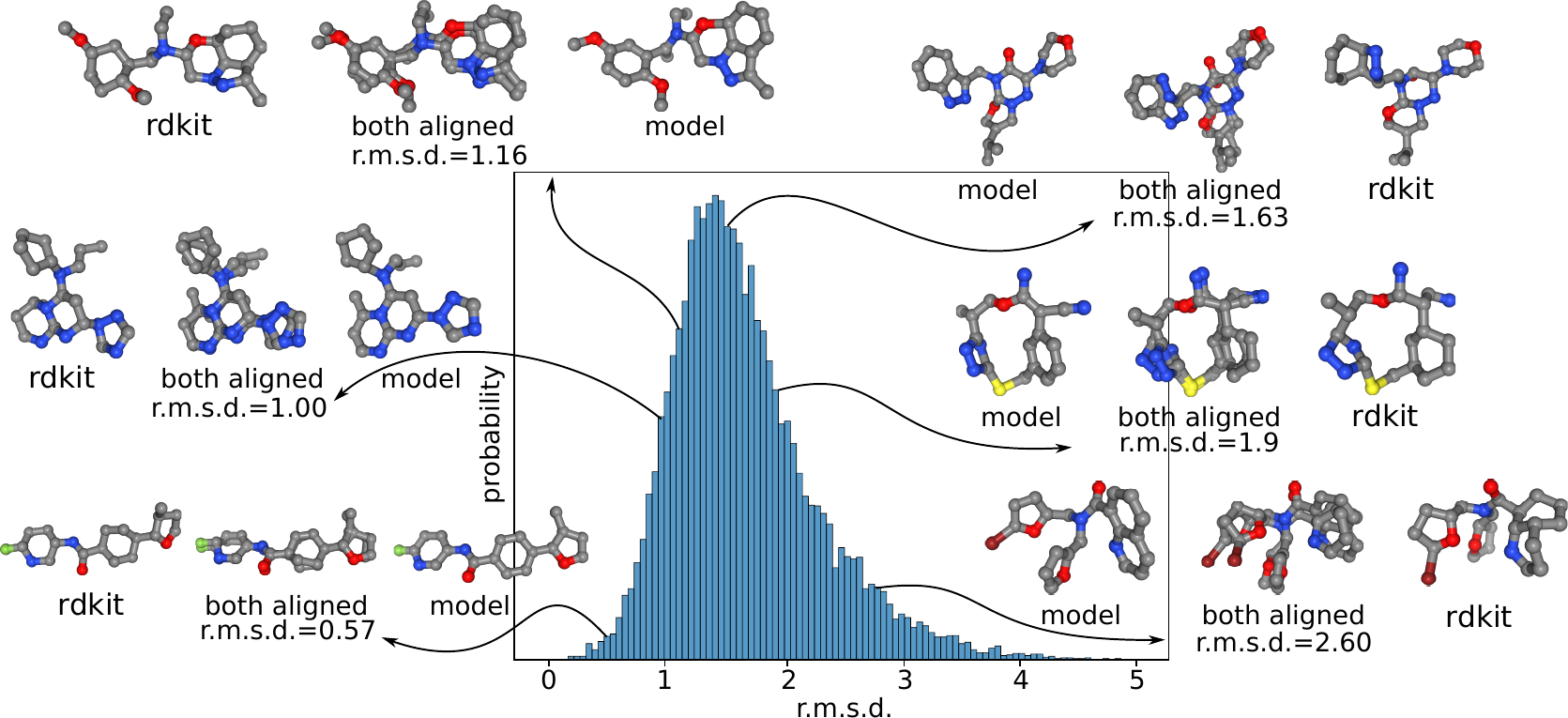}
\caption{A histogram of root mean squared deviations in atomic positions between 10K molecules sampled from the language model and their corresponding conformers generated by rdkit. Six example molecules and geometries with various r.m.s.d. values are visualized explicitly and compared with their rdkit conformers.}
\label{fig:fig2}
\end{figure*}

% discuss model and experiments and paper 
We test the ability of a transformer-based language model to generate molecules from a standard benchmark \texttt{ZINC} \citep{gomez2018automatic} using sequences parsed from the XYZ files of these molecules. We further investigate the model's ability on materials-- specifically crystals by training them to directly generate sequences parsed from CIFs (Crystallographic Information files), from two recent crystal benchmarks: \texttt{PEROV5} and \texttt{MP20} \cite{xiecrystal}. In particular, we focus on assessing the model's ability to generate valid molecules and materials that reproduce the distributional properties of the training datasets.

In addition to training language models on all datasets, we compare with state-of-art baselines models that generate molecules or materials as point clouds in 3D space \cite{gebauer2019symmetry}. Despite having no equivariance and being constrained by the data structure, to place atoms using absolute coordinates that are generated by a single character or coordinate at a time-- the results presented in this work demonstrate a language model is comparable to the ability of domain-specific 3D generative models. 

We also further demonstrate the ability of language models to generate large molecular structures in 3D. For this we show that they can scale to protein binding sites in Protein Data Bank files-- these are specific structural regions within proteins with hundreds of atoms that can only be truly represented as 3D point clouds.     

We establish that current language models are powerful generative models for chemistry that can learn to generate structural distributions of molecules, biomolecular structures, and materials-- directly in three dimensions. 

\section*{Results}

The training workflow, example molecules, materials, and a protein binding site are displayed in Figure \ref{fig:f1}. 
The model is trained to predict the next token in a sequence defined by processing chemical file formats-- XYZs, CIFs or PDBs  into sequences using different tokenization strategies. After  simplifying the file and removing unnecessary information We use two different strategies: the first is character-level tokenization (\textbf{LM-CH}), where the model must generate every necessary individual element of the file including spaces between coordinates as well as characters that indicate a newline in the file. Next, in atom+coordinate-level tokenization (\textbf{LM-AC}), the model only generates atom tokens like 'C' for carbon and coordinates tokens like '-1.98'. For each placement 4 tokens are required to place an atom in 3D space: the atom token and x,y, and z coordinate tokens. In both, we must first specify a level of numerical precision to be used and round all floating point numbers to either 1, 2, or 3 decimal places.
Additionally, because the model is not rotation and translation invariant, data augmentation by randomly rotating training structures is a useful tool to improve performance. 
More technical details about the model and training are available in the Methods section. We evaluate the model on each of the 3D chemical structure distributions detailed further in the next sections.

\subsection*{Molecules}

We test the model on sequences derived from XYZ files of molecules from the ZINC dataset \cite{irwin2005zinc} that consists of 250K commercially available molecules with on average 23 heavy atoms. We generate the XYZ files using rdkit's \cite{landrum2013rdkit} built-in conformer generation tools. While other datasets of molecules exist-- the ZINC dataset is the most established benchmark for graph and string generative models of molecules-- enabling a wider comparison. 
We train both language models on XYZ-derived sequences and first specify a numerical precision of 2 decimal places for all atomic coordinates. 

We generate 10K (thousand) molecules from the model in order to evaluate its performance and ability to sample from the distribution of molecules used in training. We evaluate the model in two ways -- the first assesses the 3D molecular geometries the model learns. Second, we compare the model using standard metrics used to assess generative models for chemistry. Samples from the model are very high quality and very similar to the training samples-- this can be seen directly by visualizing samples-- random examples are shown in the supplementary information. 

First, we assess if the language model (\textbf{LM-AC}) can learn to generate molecules with similar 3D structures that would be generated by rdkit. To do this we plot a distribution of the rdkit computed root mean square deviation (r.m.s.d.) of atomic positions between molecules generated in 3D by the language model and the corresponding molecule with 3D structure produced by rdkit's conformer tools. To attach some relative meaning to the values in the histogram-- for six different r.m.s.d. values we plot molecules generated by the language model that have different geometric structures. We also show the corresponding rdkit structure and plot in between, both molecules aligned. We label the r.m.s.d. for each as well as show which region of the histogram the molecule lies in. The model's distribution of r.m.s.d. ranges mostly between 1.0 and 2.0 although it has a heavy tail from 2.0-4.0 but quickly trails off.  We can see the model does produce molecular geometries that are close in overall structure to geometries produced by rdkit. Additional examples comparing rdkit's geometry and the language model can be found in the supplementary information.

Next, we compare molecules generated by the language model with samples from various other generative models for molecules that are widely applied. 
For these baselines, we consider models that explicitly train on 3D structures as well as models that train on molecular graph or string representations. For 3D generative models, we compare with G-Schnet \cite{gebauer2019symmetry}-- an auto-regressive 3D generative model  that places atoms using interatomic distances, equivariant normalizing flows (ENF) \cite{satorras2021n} and equivariant diffusion for 3D molecular generation (EDM) \cite{hoogeboom2022equivariant}. Additionally, we consider chemical language models using a recurrent neural network with long short-term memory \cite{hochreiter1997long} trained on either SMILES  (SM-LM) or SELFIES (SF-LM). We also train some popular deep graph generative models: junction tree variational autoencoder (JTVAE) \cite{jin2018junction} which pieces together substructures or other models that generate molecular graphs by predicting individual atoms or bonds-- these include: constrained graph variational autoencoder (CGVAE) \cite{liu2018constrained} and deep generative auto-regressive model of graphs (DGMG)\cite{Liu2018}. 

We also use standard metrics like validity, uniqueness, and novelty \cite{FlamShepherd_2021}-- to assess the model's ability to generate a diverse set of real molecules distinct from the training data. For models using graph and string representations, we use rdkit to determine validity but for 3d models we use xyz2mol \cite{jensen2020xyz2mol} to determine validity-- if can produce a valid Mol object in rdkit.  
For quantitative evaluation of any model's ability to learn its training distribution, we compute the earth mover's distance (WA) between property values of generated molecules and training molecules. We also compute the earth mover's distance between different samples of training molecules (TRAIN in Table \ref{zinc}) which acts as an oracle baseline to lower bound all earth mover distances. 
For molecular properties, we consider: 
quantitative estimate of drug-likeness (QED) \cite{bickerton2012quantifying}, 
synthetic accessibility score (SA) \cite{ertl2009estimation},  
%octanol-water partition coefficient (LogP) \cite{wildman1999prediction}, 
exact molecular weight (MW).

In Table \ref{zinc}, we can see that both language models using character and coordinate level tokenization-- achieve competitive performance 
to models using graph and string representations. Indeed, the character-level language model performs comparable to the graph models but is worse than the SMILES  (SM-LM) or SELFIES (SF-LM) language models. However, the coordinate-level language model achieves performance that is comparable to or better than all models.

\begin{table}[t]
\caption{Generation performance for ZINC.}
\label{zinc}
\scriptsize
\begin{center}
\begin{tabular}{ll|llllll}
\multirow{2}{*}{3D} & \multirow{2}{*}{Model} & \multicolumn{3}{c}{\bf Basic Metrics (\%)  $\uparrow$}  & \multicolumn{3}{c}{\bf WA Metrics $\downarrow$}  \\
& & Valid & Unique & Novel & MW & SA & QED  \\
\hline % \parbox[t]{4mm}{\multirow{6}{*}{\rotatebox[origin=c]{90}{Perov5}}}    \multirow[t]{3}{*}{Perov-5}
\parbox[t]{4mm}{\multirow{6}{*}{\rotatebox[origin=c]{90}{Not 3D }}} 
& Train   & 100.0 & 100.0 & 100.0 & 0.816 & 0.013 & 0.002 \\
& SMLM    & 98.35 & 100.0 & 100.0 & 3.640 & 0.049 & 0.005 \\
& SFLM    & 100.0 & 100.0 & 100.0 & 3.772 & 0.085 &	0.006 \\
& DGMG    & 79.63 & 100.0 & 99.38 & 88.94 & 3.163 & 0.095 \\
& JTVAE   & 100.0 & 98.56 & 100.0 & 22.63 & 0.126 & 0.023 \\
& CGVAE   & 100.0 & 100.0 & 100.0 & 45.61 & 0.426 & 0.038 \\
\hline
\parbox[t]{4mm}{\multirow{6}{*}{\rotatebox[origin=c]{90}{3D}}} 
         & ENF     & 1.05  & 96.37 & 99.72 & 168.5 & 1.886 & 0.160 \\
         & GSchNet & 1.20  &	55.96 & 98.33 & 152.7 & 1.126 &	0.185 \\
         & EDM     & 77.51 & 96.40 & 95.30 & 101.2 & 0.939 & 0.093 \\ \cline{2-8} 
  & \textbf{LM-CH}  & 90.13 & 100.0 & 100.0 & 3.912 & 2.608 & 0.077 \\
 & \textbf{LM-AC}  & \textbf{98.51} & \textbf{100.0} & {\textbf{100.0}} & {\textbf{1.811}} & \textbf{0.026} &  \textbf{0.004} \\
\hline
\end{tabular}
\end{center}
\end{table}

\subsection*{Crystals}

Next, we turn to materials like crystals which are structures that cannot be represented as graphs. Specifically, crystals are materials whose constituents atoms are arranged in a highly ordered lattice structure that extends, repeating in all directions. Crystals are stored in standard text file formats known as CIFs- Crystallographic Information Files. Within CIFs, the structural information necessary to describe the crystal includes atomic elements and coordinates as well as the parameters defining the periodic lattice. Similar to an XYZ file, CIF files include atomic elements positioned in a unit cell or lattice, with six additional parameters necessary to define the unit cell. This information can be generated before the atomic elements and positions-- either a character at a time or treating each entire parameter as a single token.  
To test if language models can generate crystals as CIF-derived sequences, we turn to curated datasets from recent work on generative models for crystals \cite{xiecrystal}. We focus on two of the datasets, the first is Perov5 \cite{castelli2012computational} which includes 18928 perovskite materials that share the same structure but differ in composition. There are 56 possible elements and all materials have exactly 5 atoms in the unit cell. The second dataset, MP20 \cite{jain2013commentary} consists of 45231 materials varying in both structure and composition. There are 
89 elements and the materials have between 1 and 20 atoms in the unit cells. We use the exact same setup and evaluation as \cite{xiecrystal}, further details regarding datasets and evaluation can be found there. 

We follow that prior work \cite{xiecrystal} and use several metrics that they used to evaluate the validity, property statistics, and diversity of generated materials. We briefly detail them here, the first is 1) Validity: a crystal is structurally valid if the shortest distance between any pair of atoms is larger than 0.5 \normalfont\AA \cite{court20203} and the composition of a crystal is valid if the overall charge is neutral as computed by \texttt{SMACT} \citep{davies2019smact}. 2) Coverage (Cov) COV-R (Recall) and COV-P (Precision) \cite{xu2021learning}, measures the similarity between ensembles of generated materials and ground truth test materials.  3) Property statistics, we also compute the earth mover's distance (WA) between the property distribution of generated materials and test materials. For properties, we use density ($\rho$) and number of unique elements (\# elem.). Following \cite{xiecrystal} we sample 10K materials after training to compute evaluation metrics. 

We compare the language model with the baselines taken from \cite{xiecrystal}, which include the latest state-of-the-art generative models and methods. These include: FTCP ~\citep{ren2020inverse} is a 1D CNN-VAE trained over a crystal representation that concatenates various properties (atom positions, atom types, diffraction pattern, etc). GSchNet was also compared with in \cite{xiecrystal} first computationally determining lattices afterward, then using a modified version (PGSchNet) that directly incorporates periodicity. Lastly, we also compare with the best-performing model in \cite{xiecrystal}, the crystal diffusion variational autoencoder (CDVAE). We also include an oracle (TRAIN) that defines an upper bound for validity and coverage and a lower bound for property statistics-- computed using a sample from the training data. The results are displayed in Table \ref{crystals}.  

\begin{table}[t]
\caption{Crystal generation performance.}
\label{crystals}
\scriptsize
\begin{center}
\begin{tabular}{ll|llllll}
\multirow{2}{*}{Data} & \multirow{2}{*}{Model} & \multicolumn{2}{c}{\bf Valid (\%)  $\uparrow$}  & \multicolumn{2}{c}{\bf COV (\%) $\uparrow$}  & \multicolumn{2}{c}{\bf WA $\downarrow$}  \\
& & Struc. & Comp. & R. & P. & $\rho$ & \#  \\
\hline % \parbox[t]{4mm}{\multirow{6}{*}{\rotatebox[origin=c]{90}{Perov5}}}    \multirow[t]{3}{*}{Perov-5}
\parbox[t]{4mm}{\multirow{6}{*}{\rotatebox[origin=c]{90}{Perov5}}} & Train & \textbf{\emph{100.0}} & \textbf{\emph{98.60}} & \textbf{\emph{100.0}} & \textbf{\emph{100.0}}  &\textbf{\emph{0.010}} & \textbf{\emph{0.008}} \\
& FTCP     & 0.24 & 54.24 & {0.00} & {0.00} & 10.27 & 0.630 \\
& GSchNet  & 99.92 & \textbf{98.79} & {0.18} & {0.23} & 1.625  &	0.037 \\
& PGSchNet & {79.63} & {\textbf{99.13}} & {0.37} & {0.25} & {0.276}  & {0.455} \\
& CDVAE    & \textbf{100.0} & \textbf{98.59} & {\textbf{99.45}} & 98.46 & 0.126 &  0.063 \\ \hline
& \textbf{LM-CH}     & \textbf{100.0} & \textbf{98.51} & {\textbf{99.60}} & {\textbf{99.42}} & \textbf{0.071} &  \textbf{0.036} \\
& \textbf{LM-AC}     & \textbf{100.0} & \textbf{98.79} & {\textbf{98.78}} & {\textbf{99.36}} & \textbf{0.089} &  \textbf{0.028} \\
\hline \hline
\parbox[t]{4mm}{\multirow{6}{*}{\rotatebox[origin=c]{90}{MP20}}} & Train & \textbf{\emph{100.0}} & \textbf{\emph{91.13}} & \textbf{\emph{100.0}} & \textbf{\emph{100.0}} & \textbf{\emph{0.051}} &  \textbf{\emph{0.016}} \\
 & FTCP & 1.55 & 48.37 & {4.72} & {0.09} & 23.71 &  0.736 \\
 & GSchNet & 99.65 &	75.96 & {38.33} & {\textbf{99.57}} & 3.034 &	0.641	\\
 & PGSchNet & {77.51} & {76.40} & {41.93} & {\textbf{99.74}} & {4.04} & 0.623 \\
 & CDVAE & \textbf{100.0} & 86.70 & 99.15 & {\textbf{99.49}} & \textbf{0.688} & 1.432  \\ \cline{2-8}
 & \textbf{LM-CH}   & 84.81 & 83.55 & \textbf{99.25} & 97.89 & 0.864 &  \textbf{0.132} \\
 & \textbf{LM-AC}   & 95.81 & \textbf{88.87} & {\textbf{99.60}} & 98.55 & 0.696 &  \textbf{0.092} \\
\hline
\end{tabular}
\end{center}
\end{table}

We train language models on CIF-derived sequences and first specify a numerical precision of 3 decimal places for all floating-point numbers (unit cell parameters and coordinates) in the CIF files. 

From the results, it is clear that language models are capable of generating novel materials that maintain the properties of the crystals in both training distributions. 
Both character and coordinate level language models show strong performance in all evaluation metrics from validity, property statistics, and diversity. Indeed in the smaller crystal dataset PEROV5, language models achieve better metrics over the baseline models. In the larger, more structurally diverse dataset MP20, the coordinate-level language model achieves the best performance in 3 of six metrics but is close to state-of-the-art performance in the other metrics as well.  
The character-level language model is slightly worse but still has comparable performance to the other baselines including the CDVAE. 

There are more complex materials to test the capabilities of language models on but the experiments on these materials and the results indicate the strong potential of language models for materials generation and design. It is important to note that, beyond these metrics, more work is necessary to verify the results with computational simulation and experiments.

\begin{figure*}[tbhp]
\centering
\includegraphics[width=.99\linewidth]{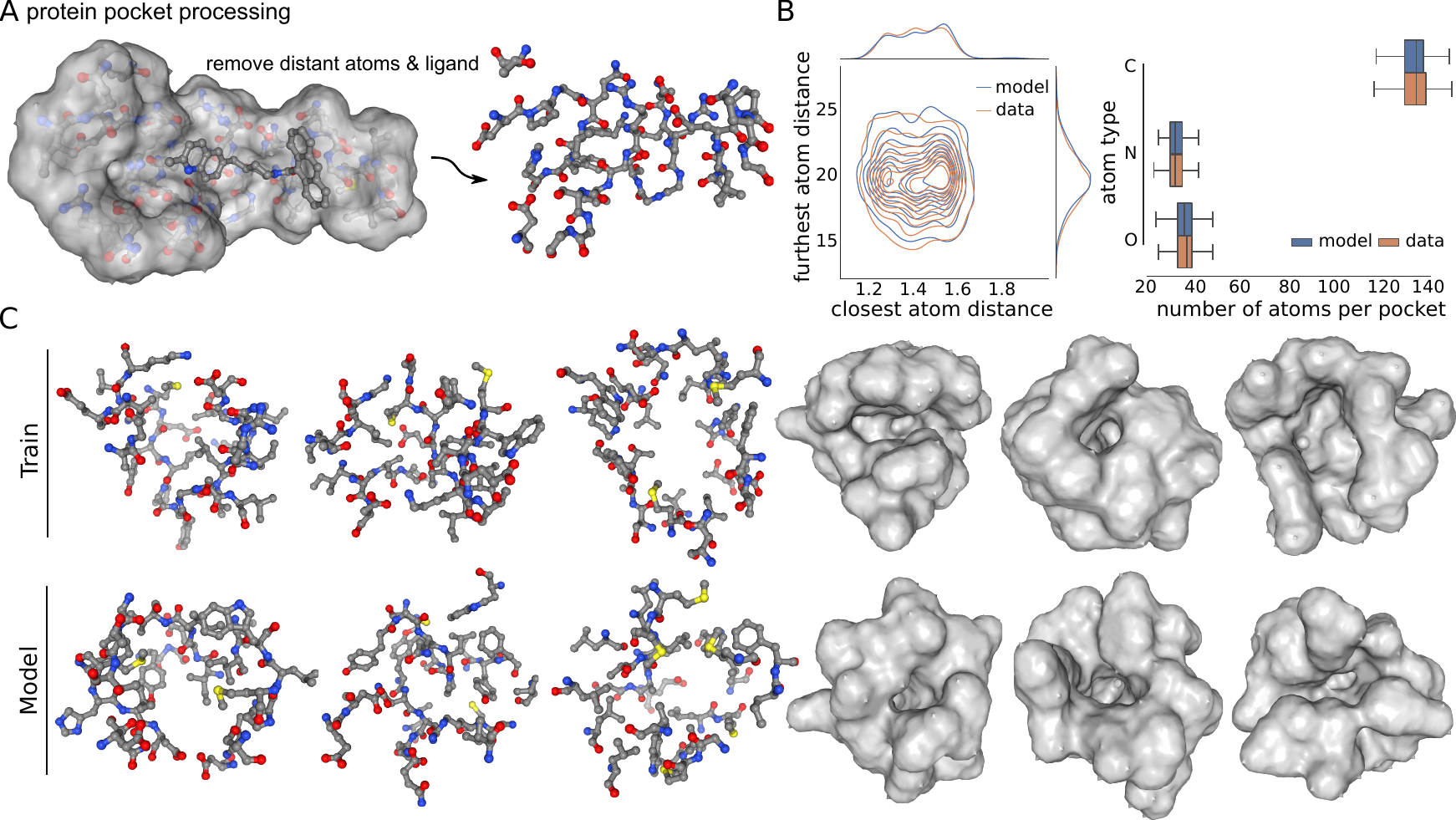}
\caption{A) Protein pockets are pre-processed by removing residues far from the center of the pocket-ligand complex. B) A comparison between the model and training data distribution of interatomic distance between 10 random pocket atoms and the closest and furthest pocket atoms. Additionally, we show a box plot for the number of carbon, nitrogen, and oxygen C) Six different examples from the training data and sampled from the language model the first 3 are plotted showing individual atoms, and the last three show the surface of the pocket.}
\label{pockets}
\end{figure*}

\subsection*{Protein Binding Sites}

For the most challenging task, we test if language models can generate biomolecular structures that are stored within PDB files. To test this, in a limited way, we train language models on sequences derived from PDB files storing protein binding sites. These are regions on proteins that bind to ligands-- other molecules including small molecules, peptides, or other proteins. Typically they are small subsections of a protein containing a few dozen residues and hundreds of atoms that define a distinct geometric pocket or cavity.  

We use a dataset of $\sim$180K protein-ligand pairs from \cite{luo20213d}. As shown in Figure \ref{pockets} A) we process the protein binding sites by removing all atoms in residues that are furthest from the center of the protein-ligand complex until there are roughly 200-250 atoms remaining. The training structures are just the remaining residues-- the ligand is removed as well. No generative model based on graphs would be able to generate the protein pockets directly-- their 3D structure is what gives rise to their function.  

Similar to XYZ files-- PDB files have atom information like elements and coordinates but have additional information related to protein structure such as every atom's residue. Therefore, after simplifying and removing other extraneous information, we convert the files to sequences using residue information as well-- which we jointly tokenize with atom information. For example, atoms that are part of cysteine residues can be identified with one of the following tokens: $\texttt{CYS-C}, \texttt{CYS-N}, \texttt{CYS-O}, \texttt{CYS-S}$. This will allow the model to organize how each atom is placed by associating it with a local neighborhood defined by its specific amino acid. 

We force the numerical precision of each atomic position to two decimal points and identical to atom+coordinate tokenization we use a single token for each x,y, and z coordinate of the atomic position entirely so that every atom has four tokens associated with it: one to identify its atomic element and residue as well as three tokens for its atomic position. Character-level tokenization produces sequences that are substantially longer so we do not experiment with it in this task. 

 We cannot use  xyz2mol \cite{jensen2020xyz2mol} to assess validity in this context since is only applicable to smaller molecules, similarly, other standard and distribution metrics for drug-like molecules are not meaningful for protein pockets. Instead, to measure validity, after sampling 10K pockets from the model, we check each residue individually with xyz2mol, and make sure the atom composition of each residue is correct. Additionally, to check if there are any overlapping atoms among residues we check if any atoms from different residues are closer than the smallest possible bond distance. Almost all pockets, or $\sim$99\% of pockets sample pass the xyz2mol \& residue check while $\sim$5\% of pockets fail the overlap check- we show some examples of these pockets in the supplementary information. 

We also compare the training distribution to the model's learned distribution: first, using a bivariate kernel density estimate, we plot the joint distribution of the interatomic distance between pocket atoms and their furthest or closest neighbor. In Figure \ref{pockets} B) we can see the model closely matches the training distribution for the closest and furthest neighboring atoms. In addition, in Figure \ref{pockets} B), pockets sampled from the language model and training pockets have a similar number of carbon, nitrogen, and oxygen atoms. 

To conduct a sanity check to see if the model is memorizing, we check the ordering of the residues essentially the amino acid sequence of the binding site (defined by the order in which residues appear in the PDB sequence ignoring coordinates ie $\texttt{ARG-SER-ASP-ILE}\cdots$) in pockets generated by the model. For comparison-- out of the $\sim$180K training pockets approximately 86.1 \% have unique orderings. Similarly, evaluating the pockets that the language model generates, we get $\sim$89.8.4\% unique orderings of residues and further, of these pockets, 83.6\% have novel residues orderings that do not occur in the training pockets. This indicates the model is learning to generate mostly novel protein pockets with new amino acid sequences while maintaining the higher-level geometric structure that defines a protein pocket.   

Additionally, we display a few examples of training and model-generated pockets in Figure \ref{pockets} C) including both pockets showing individual residues and atoms as well as pockets with the surface explicitly rendered-- which helps highlight the actual geometric structure of the pocket. Qualitatively, both ways of visualizing the pocket, demonstrate that the language model generates pockets that do have a similar geometric structure to the training examples. 

% Authors should submit SI as a single separate SI Appendix PDF file, combining all text, figures, tables, movie legends, and SI references. SI will be published as provided by the authors; it will not be edited or composed. Additional details can be found in the \href{https://www.pnas.org/authors/submitting-your-manuscript#manuscript-formatting-guidelines}{PNAS Author Center}. The PNAS Overleaf SI template can be found \href{https://www.overleaf.com/latex/templates/pnas-template-for-supplementary-information/wqfsfqwyjtsd}{here}. Refer to the SI Appendix in the manuscript at an appropriate point in the text. Number supporting figures and tables starting with S1, S2, etc.

%%%%%%%%%%%%%%%%%%%%%%%%%%%%%%%%%%%%
\section*{Discussion}
%%%%%%%%%%%%%%%%%%%%%%%%%%%%%%%%%%%%

We have demonstrated that language models can learn to generate molecules, materials, and biomolecular structures directly in three dimensions when trained successfully on sequences derived from chemical file formats like XYZ, CIF, and PDB. The results show that language models are powerful generative models capable of learning to generate complex chemical structures in three dimensions. Language models are not just limited to simple string molecular representations like SMILES and SELFIES but can directly learn structured representations by merely predicting the next token in sequences derived from these representations. In contrast to most generative models for molecules, materials, and biomolecules that are designed for very specific classes of molecules -- we demonstrate that language models, without any architecture changes and simply using next-token prediction can generate a wide variety of different chemical structures. We showed that character-level language models were able to model small drug-like molecules and simple crystals. Even further with atom and coordinate-level tokenization, language models can generate biomolecular structures like protein binding sites that contain hundreds of atoms.  

In future work, building on these results, there is enormous potential to use language models for inverse design of molecules or materials to optimize properties that depend on the geometric structure. Additionally, we are interested in testing language models in other more complex classes of molecules and materials like metal-organic frameworks and other structures in molecular biology. Another important potential area is structure-based drug discovery. Other aspects should be explored for further success including the use of different tokenization strategies. A particular issue when directly tokenizing entire coordinates is the size of the vocabulary-- which will grow enormously as the structure of the molecule or material being modeled grows. Predicting absolute coordinates which are not rotation or translation invariant is challenging for structures with hundreds of atoms-- training on even larger structures will be difficult and may require large amounts of data. 

We predict that larger and larger datasets of molecules and materials will become available in the future. As more and more data becomes available-- language models will continue to improve and demonstrate their power by modeling tasks once thought impossible for them.

\section{Methods}

\subsection{Chemical structure Representations}

\paragraph{Molecules (XYZ files)}

We represent a molecule as a point cloud of $n$ atoms with elements $e_i \in \{\texttt{C},,\dots \}$ and positions $ x_i,y_i,z_i \in \mathbb{R}$-- as follows
\begin{equation}
\mathcal{M}=  (e_1, x_1, y_1, z_1, \dots , e_n, x_n, y_n, z_n )      
\end{equation}

\paragraph{Crystals (CIF files)}

Any crystal can be represented similarly but must include necessary information about the unit cell or lattice in addition to atomic positions and elements. The unit cell is a parallelepiped, so there are six necessary lattice parameters taken as the lengths of the cell edges ($\ell _a, \ell_b, \ell_c$) and the angles between them ($\alpha, \beta, \gamma $). The positions of atoms inside the unit cell are described by fractional coordinates ($x_i, y_i, z_i$) along the cell edges. Thus crystals can be completely described using the following tuple: 
\begin{equation}
\mathcal{C}=  (\ell _a, \ell_b, \ell_c, \alpha, \beta, \gamma, e_1, x_1, y_1, z_1, \dots , e_n, x_n, y_n, z_n )      
\end{equation}

\paragraph{Protein pockets (PDB files)}

We represent a protein pocket as a point cloud of $n$ atoms with residue-atom indicators $a_i \in \{\texttt{HIS-C},\texttt{HIS-N},\dots \}$ and positions $ x_i,y_i,z_i \in \mathbb{R}$-- as follows
\begin{equation}
\mathcal{P}=  (a_1, x_1, y_1, z_1, \dots , a_n, x_n, y_n, z_n )      
\end{equation}

\subsection{Tokenization}
Ignoring special tokens, character-level models use a small vocabulary of $\sim$30 tokens consisting of atom type tokens $ \texttt{C},\texttt{N}, \dots $ digit characters and minus sign $ \texttt{1-9},\texttt{-}  $, and other file symbols like newline character which we represent with a hashtag as well as an empty space token $ \texttt{' '} $. 

Atom+coordinate-level models use a larger vocabulary of $\sim$100-10K tokens consisting of atom types tokens $ \texttt{C},\texttt{N}, \dots $ or atom-residue tokens $\{\texttt{HIS-C},\texttt{HIS-N},\dots$ and coordinate tokens like $ \texttt{-1.9}, \texttt{-1.98}$ or $\texttt{-1.987} $-- these range from the smallest to largest coordinate values and can have restricted precision between 1 and 3 decimal places. 

\subsection{Language Modeling} 

we frame language modeling as unsupervised distribution estimation from a set of examples $(x_1,x_2,\dots ,x_n)$ each composed of variable length sequences of tokens $t_i$ such that $x=(t_1, t_2,\dots , t_n)$. The sequences here are chemical structures so have many possible orderings (restricted by file and structural information) but regardless we factorize the joint probabilities over
\begin{align}\label{eq:mol}
    p(x)=\prod _{i=1}^n p(t_n | t_{n-1},\dots t_1)
\end{align}
this probability is modeled using a Transformer \cite{radford2018improving} with parameters that are trained using stochastic gradient descent. 
 
 \subsection{Data Augmentation}

Since the model is not invariant to rotations or translations, to improve performance at every epoch we can randomly rotate any training structure about its center to expand the training data. Models trained without data augmentation can still achieve performance close to SOTA. 

\subsection{Model architecture and Training}

We use Transformers with GPT architecture \cite{radford2018improving} that have roughly between $\sim$1 and 100 million parameters and use 12 layers, embedding size between $\{128,1024\}$, and 4 to 12 attention heads. For training we use a small batch size between [4,32] structures, and a starting learning rate between $[10^{-4},10^{-5})$, that is decayed to $9\cdot 10^{-6}$ over training. Example code can be found at
\href{https://github.com/danielflamshep/xyztransformer}{https://github.com/danielflamshep/xyztransformer}.

\section{Acknowledgements}   A.A.-G. acknowledge funding from Dr. Anders G. Fr{\o}seth.
A.A.-G. also acknowledges support from the Canada 150 Research Chairs Program, the Canada Industrial Research Chair Program, and from Google, Inc.
Models were trained using the Canada Computing Systems \cite{baldwin2012compute}. A.A.-G. also acknowledges support from the Acceleration Consortium at the University of Toronto.

%%%%%%%%%%%%%%%%%%%%
%%% BIBLIOGRAPHY %%%
%%%%%%%%%%%%%%%%%%%%
\bibliographystyle{apsrev4-1}
\bibliography{main}
\clearpage

%%%%%%%%%% Merge with supplemental materials %%%%%%%%%%
\pagebreak
%%%%%%%%%%%%%%%
\begin{widetext}
%%%%%%%%%%%%%%%%

\newpage

\section*{Supplementary}

%%%%%%%%%%%%%%%%%%%%%%%%%%%%%%%%
\setcounter{equation}{0}
\setcounter{figure}{0}
\setcounter{table}{0}
\makeatletter
\renewcommand{\theequation}{S\arabic{equation}}
\renewcommand{\thefigure}{S\arabic{figure}}

\begin{figure}[ht]
\centering
\includegraphics[width=0.9\textwidth]{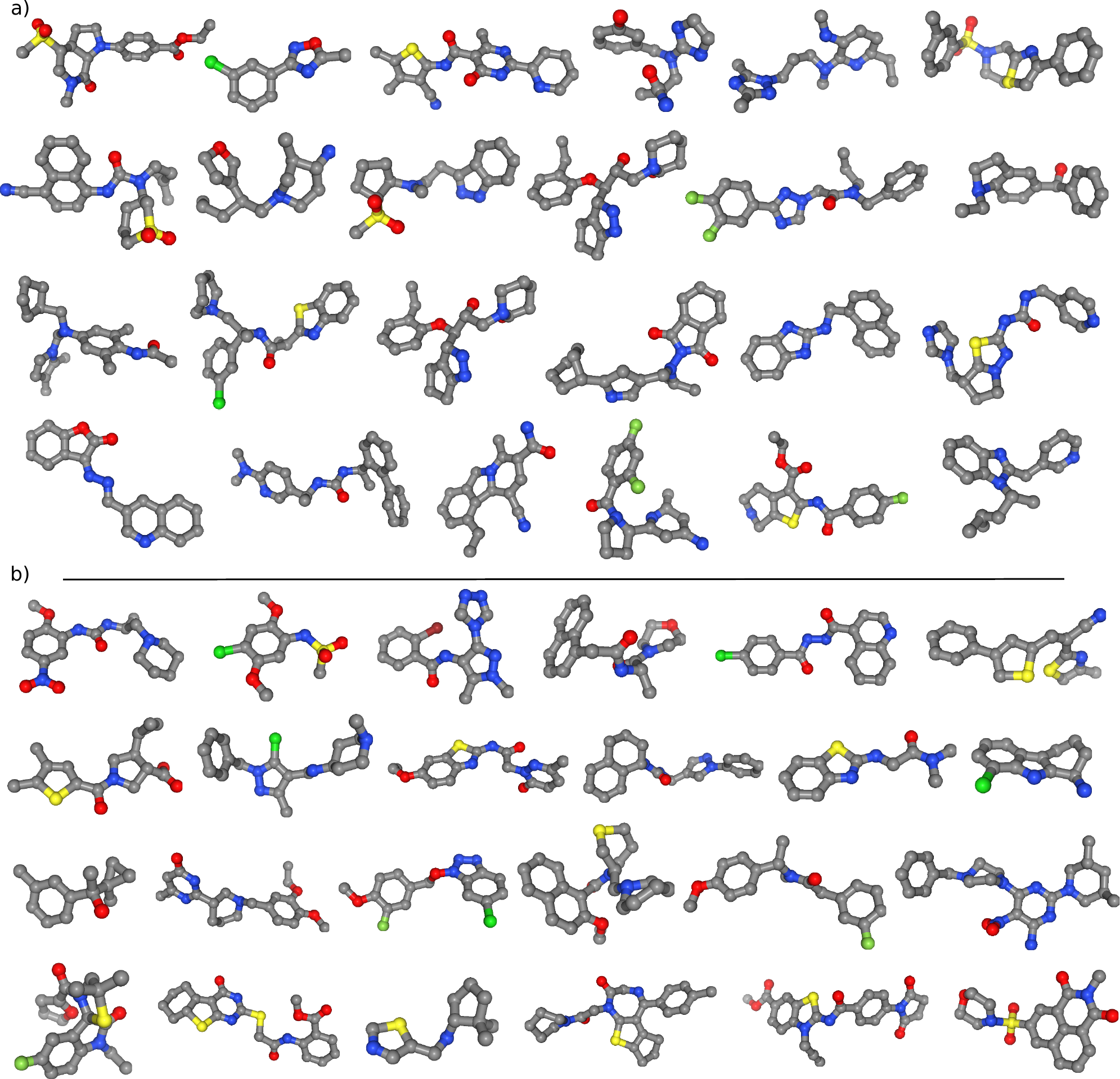}
\caption{a) Examples of training molecules in three dimensions. b) Samples of molecules generated by the language model.}
\end{figure}

\begin{figure}[ht]
\centering
\includegraphics[width=0.8\textwidth]{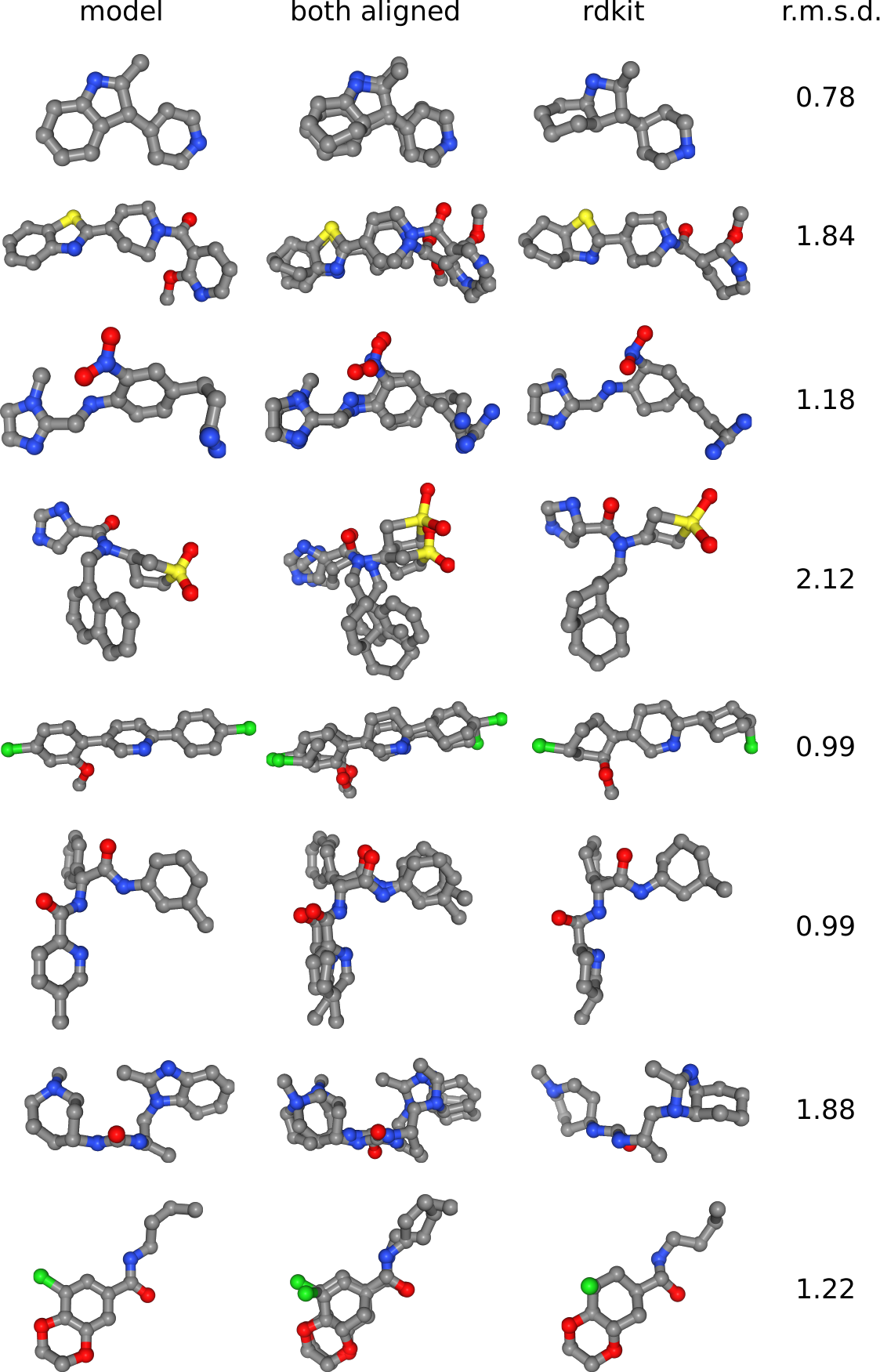}
\caption{example molecules and geometries with various r.m.s.d. values are visualized explicitly and compared with their rdkit conformers.}
\end{figure}

\begin{figure}
\centering
\includegraphics[width=0.9\textwidth]{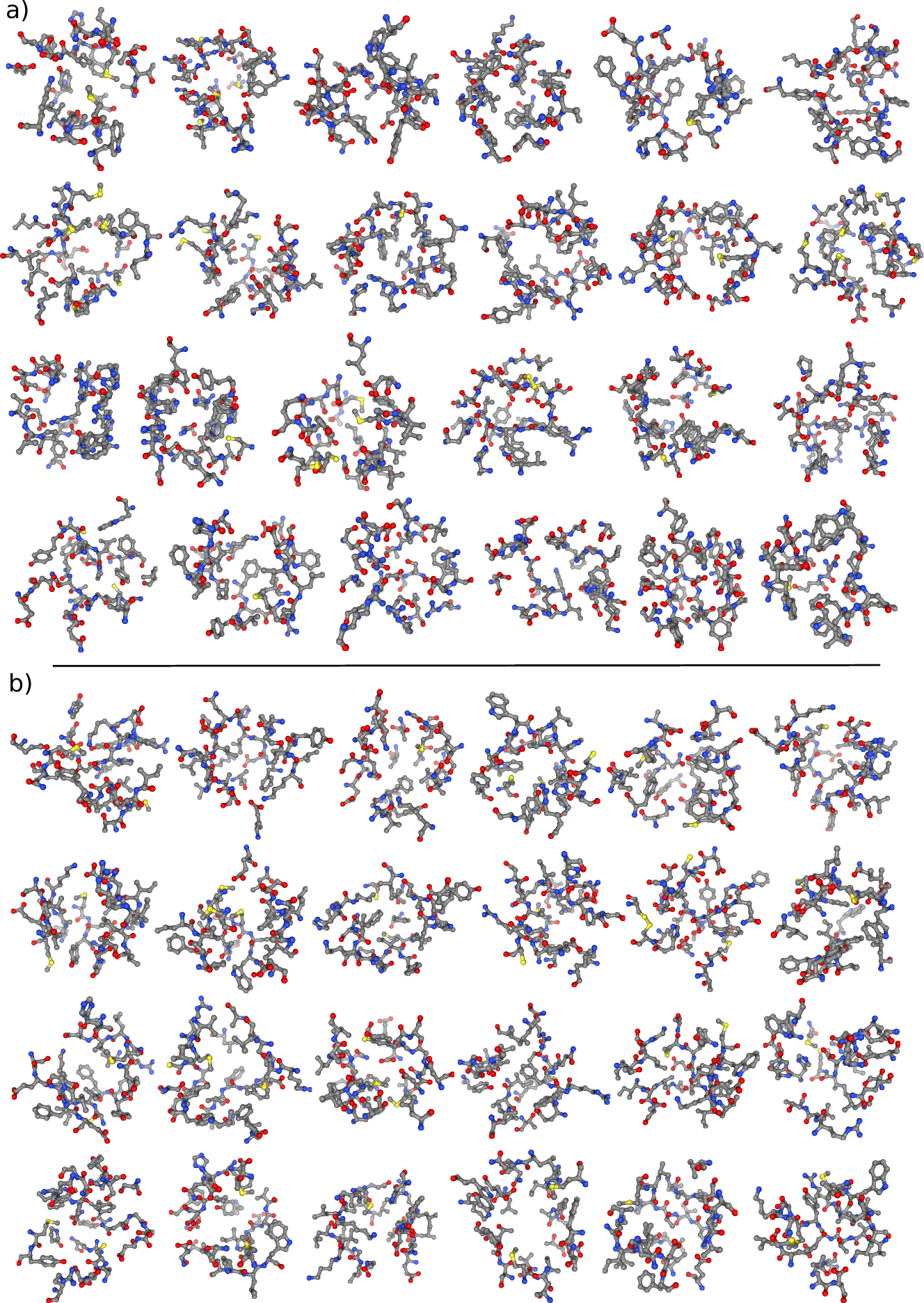}
\caption{a) Examples of training protein pockets. b) Samples of protein pockets generated by the model.}
\end{figure}

\begin{figure}
\centering
\includegraphics[width=0.9\textwidth]{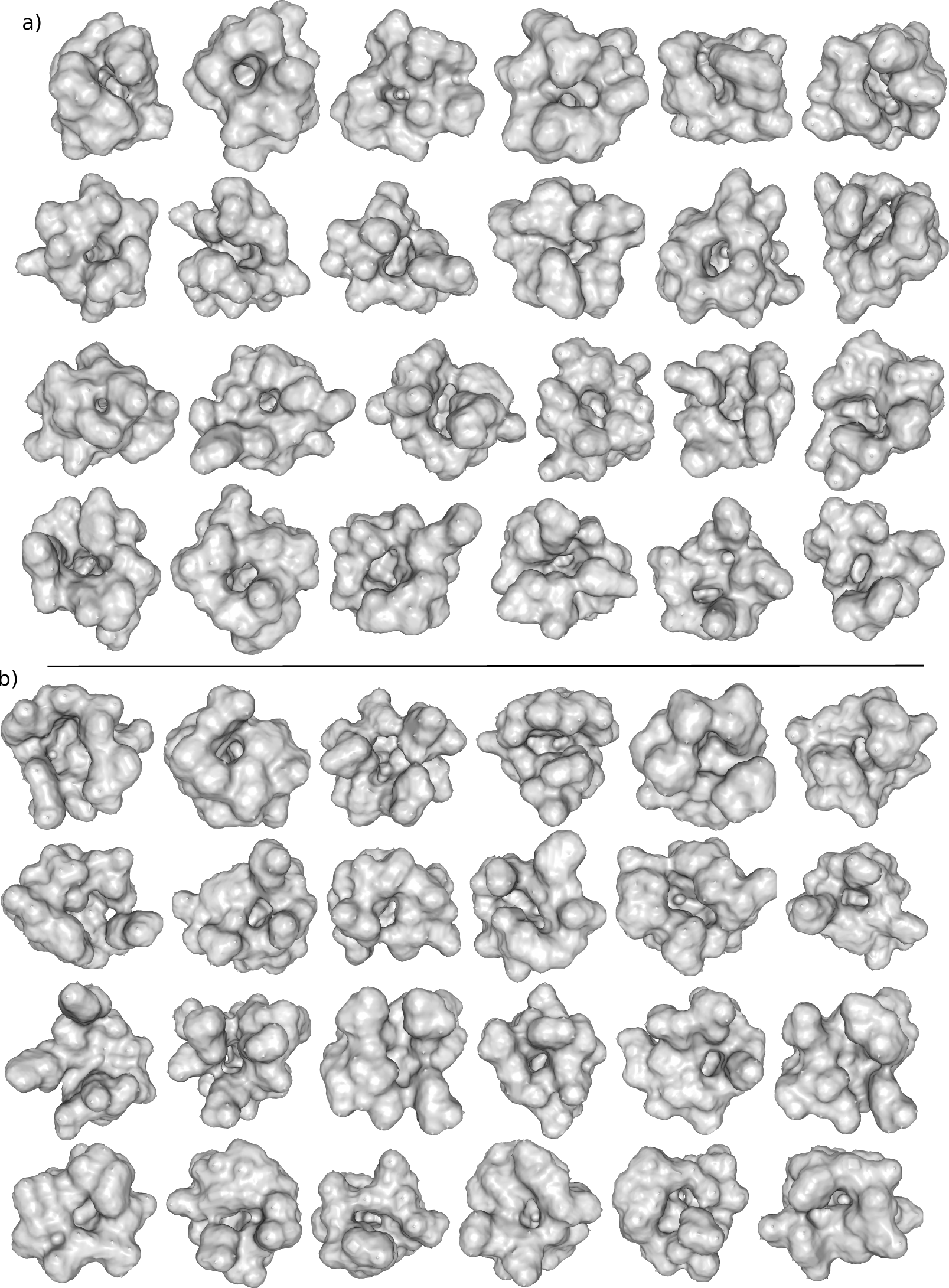}
\caption{a) Examples of training protein pockets with their surface rendered. b) Samples of protein pockets generated by the language model.}
\end{figure}

\begin{figure}
\centering
\includegraphics[width=0.9\textwidth]{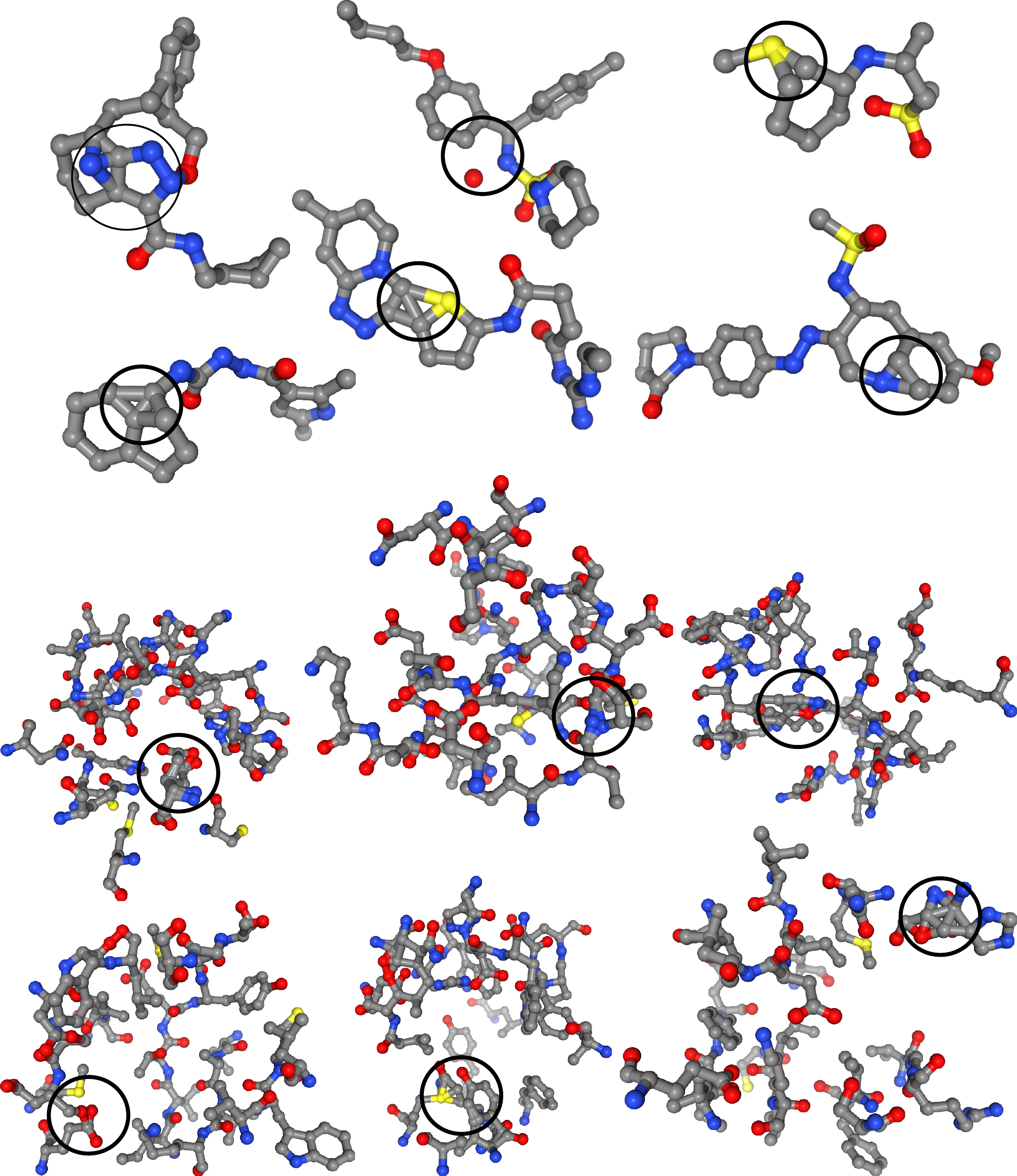}
\caption{Examples of generated structures that are invalid.}
\end{figure}

%%%%%%%%%%%%%%%%
\end{widetext}
%%%%%%%%%%%%%%%%%%%%%

\end{document}